\documentclass[11pt,a4paper]{article}
\usepackage[hyperref]{emnlp2020}
\usepackage{times}
\usepackage{latexsym}
\usepackage{graphicx}  
\usepackage{color}
\usepackage{amsmath}
\usepackage{amssymb}
\usepackage{algorithm}
\usepackage{pgfplots}
\usepackage[noend]{algpseudocode}
\usepackage{booktabs}
\usepackage{multirow}
\usepackage{booktabs}
\definecolor{orange}{RGB}{255,127,0}

\usepackage{microtype}

\aclfinalcopy 
\def\aclpaperid{288} 

\title{
Template Controllable \textit{keywords-to-text} Generation}
\author{
    Abhijit Mishra\thanks{Currently works at Apple inc.}, Md. Faisal Mahbub Chowdhury, Sagar Manohar, \\
    \bf Dan Gutfreund \and Karthik Sankaranarayanan \\
  IBM Research \\
  {\tt abhijitmishra.530@gmail.com}\\
  {\tt kartsank@in.ibm.com}\\
  {\tt \{mchowdh, dgutfre\} @us.ibm.com}\\
  {\tt Sagar.Mohan.Manohar@ibm.com}
}
\date{}

\begin{document}
\maketitle
\begin{abstract}

This paper proposes a novel neural model for the understudied task of generating text from keywords. The model takes as input a set of \textbf{un-ordered keywords}, and \textbf{part-of-speech (POS)} based template instructions. This makes it ideal for \emph{surface realization} in any NLG setup. The framework is based on the \emph{encode-attend-decode} paradigm, where keywords and templates are encoded first, and the decoder judiciously attends over the contexts derived from the encoded keywords and templates to generate the sentences. Training exploits \emph{weak supervision}, as the model trains on a large amount of labeled data with keywords and POS based templates prepared through completely \emph{automatic} means. Qualitative and quantitative performance analyses on publicly available test-data in various domains reveal our system’s superiority over baselines, built using state-of-the-art neural machine translation and controllable transfer techniques. Our  approach is indifferent to the order of input keywords. 

\end{abstract}
\section{Introduction}
\label{sec:intro}
The problem area of \emph{data-to-text} generation has seen a lot of interest in the language generation community recently \cite{gatt2018survey,castro-ferreira-etal-2019-neural,puduppully-etal-2019-data,ma-etal-2019-key,chen-etal-2019-enhancing,gong-etal-2019-enhanced}, primarily because it has several real world applications such as query completion, story generation, report generation, dialogue response generation for virtual assistants, support systems for second language writing, and many more. Further, with the advent of ``data-hungry'' neural models, such data-to-text NLG systems provide mechanisms for synthetic data preparation, data-augmentation, and adversarial example generation, to advance core model development.

A key challenge in data-to-text NLG is \emph{surface realization} of content \textit{i.e.,} constructing fluent sentences from input data, often available as lists of keywords.  This paper presents a simple but effective solution to this crucial problem. Note that, tackling different input orders of the keywords is of utmost importance here; keywords can appear from a structured source such as database tables or knowledge graphs in any order. 

Learning language generators has always been quite challenging,  primarily because of the higher combinatorial complexity of the output space. A workaround is to train supervised generators that reduce this complexity through supervision. Popular examples include machine translation \cite{BahdanauCB14}, image captioning \cite{VinyalsTBE15}, paraphrase generation of varying degree of language complexity \cite{iyyer-etal-2018-adversarial, li-etal-2018-paraphrase,mishra-etal-2019-modular,surya-etal-2019-unsupervised} and creative text generation \cite{Yu:aaai:2017, zhang-lapata-2014-chinese}. A clear desiderata for such systems is bulky and often expensive parallel corpora of input-output instances.

Direct supervision works in settings where the contract between the input and output is simple and clear\footnote{\textit{e.g.,} in Machine Translation it is safe to focus on learning mappings between every unique input to one possible outcome.}, and thus, learning the mapping can be based on paired corpora of well-defined input-output forms. However, for \emph{surface realization from keywords}, a supervised solution based on a paired corpora of keywords and sentences may not be adequate. For example, a keyword set of \textit{(`victim', `vanessa', `demons')} can be translated to diverse natural language forms following different lexical and syntactic choices. This may require altering the input order and also in some cases, hallucinating \textit{function-words} (\textit{e.g.,}``Vanessa can also become a victim of demons''). And, simply training the systems with keyword and sentence pairs may restrict their ability to generalize beyond language structures that are seen in the training data.

We propose a system that performs  syntax-driven language generation from keywords by considering templates as additional input. As templates, we use POS tag sequences automatically derived from  a large number of unlabelled sentences. POS-based templates are easy to interpret as well as inexpensive to obtain, during both training and runtime.  We demonstrate that considering POS-based templates has several other advantages:
\begin{itemize}
    \item It makes the system agnostic to the order in which keywords appear in the input, which is an important requirement for keywords-to-text generation.
    \item Keywords by themselves only provide topical information and do not contain information regarding the lexical and syntactic choices the generator has to make;  POS-based templates help overcome this problem.
    \item The end-user has a better control over the generation process; the user can easily specify a template of her/his choice at run-time, say, by choosing a sentence from a of a list of exemplars whose POS sequence can be automatically derived. 
\end{itemize}


Our framework is based on the \emph{encode-attend-decode} paradigm, where keywords and templates are encoded first using linear and recurrent units. The decoder carefully attends over the contexts derived from the encoded keywords and templates. Words are then produced by either \texttt{(i)} generating \textit{morpho-syntactic} and \textit{semantic} variations of the input keywords, such as inflected forms and synonyms, or \texttt{(ii)} inferring suitable \textit{function-words} from the vocabulary. For training, the system relies on automatically tagged POS-based templates and keywords. The keywords are comprised of noun, verb, adjectives and adverbs, and are automatically identified through the POS tags of the corresponding sentences from a large volume of unlabeled data (see Section \ref{sec:dataset}). 

Qualitative and quantitative performance analyses on the publicly available benchmark data \cite{chen-etal-2019-controllable} reveal our system’s superiority over baselines. The analyses also show that our system can tackle subtleties in data-to-text generation such as \texttt{(a)} diverse linguistic structures and styles, \texttt{(b)} inadequate/spurious content in the input, and \texttt{(c)} changes in input keyword order. We will release the code and data for academic use.

\section{Related Work}
\label{sec:related}
The specific problem of text generation from keywords has not received much attention. \newcite{uchimoto-etal-2002-text} first proposed a system which uses \emph{n-grams} and dependency trees to generate sentences. The system was built for Japanese. Recently, \newcite{Song:8784475} have exploited recurrent neural networks to generate the context before and after a single input keyword in Chinese.

Controlling text generation through auxiliary inputs has  received interest mainly in \textit{text-to-text} domain  \cite{kabbara-cheung-2016-stylistic}. The controllable \textit{plug-n-play language model} by \newcite{dathathri2019plug} is such a recently proposed approach. In their method, while, the generator is able to produce a fluent output based on the control specification, the generation process is still open-ended and may not adhere to any user-desired syntax. \newcite{Hu:2017:TCG:3305381.3305545} create a variational auto-encoder  (VAE) framework which provides minimal control options like flipping sentiment and flipping tense.  The system does not accept templates and is limited to generating only short text. \newcite{ghosh-etal-2017-affect} describe a method to customize the degree of emotional content in generated sentences. This system also cannot accept templates, has a fixed set of categories of emotions, and crucially, relies on actual textual data annotated with these categories. A similar effort by \newcite{ficler-goldberg-2017-controlling} attempts to control the linguistic properties of the text through a language model conditioned on a particular style. They operate in the movie-review  domain, where the possible styles are limited to \textit{theme, sentiment, professional, descriptive} \textit{etc.} These styles can take a limited set of values which the generated text should conform to, and the system lacks data transformation ability.
\newcite{jhamtani-etal-2017-shakespearizing} explore an approach to apply Shakespearean English style to modern English texts. The model uses an external dictionary of stylistic words and uses that for carrying out word replacement by copying the style; this may not always retain the desired meaning and coherence. 


Regarding template controlled generation, \newcite{iyyer-etal-2018-adversarial} propose \textit{syntactically controlled paraphrase network} (SCPN), a way to transform an input sentence based on templates given in the form of ``parse'' trees that are recurring in a language. The system can not transform input in the form of data (represented in keywords), and rather relies on well formed sentences and corresponding complete parse trees. It is worth noting that, though the system can accept an input template (such as \texttt{(S(NP)(ADVP)(VP))}, these templates are syntactically rigid and also hard to interpret. 

\newcite{chen-etal-2019-controllable} propose an approach which uses a sentence as a \textit{syntactic exemplar} rather than requiring an external parser. The authors benchmark their system against SCPN and showed competitive performance. However, this system is not designed to accept data/keywords as input (unlike our system) which can take up keywords in any order. We employ this system as a baseline for comparison. 

Recently, \newcite{Wang-abs-1901-09501}, inspired by the data-to-text generation dataset \cite{wiseman-etal-2017-challenges}, describe a method to generate sentence given a structured record (e.g. \textit{\{PLAYER: Lebron, POINTS: 20, ASSISTS: 10\}}), and a reference sentence (e.g. \textit{Kobe easily dropped 30 points}). Theirs is a different task and involves manipulating the reference text (by rewriting/adding/deleting text portions) to ensure fidelity with respect to the structured content. In our case, the keywords are not structured or even ordered and may require morphological, and syntactic transformations (change in number, tense, aspect); hence, rewriting/adding/deleting of text portions is not feasible.

\newcite{laha-2019-cl}, in similar ways as above, propose a modular system that convert entries in structured data (in tabular format) to canonical form, generate simple sentences from the canonical data, and, finally, combine the sentences to produce a coherent and fluent paragraph description. Their approach assumes table row representations as a collection of binary relations (or triples), which is a different task-setup than ours.

To the best of our knowledge, systems for translating  order invariant keywords to elaborate natural language text remain elusive.
\begin{figure*}[t]
\centering 
\includegraphics[scale=0.51]{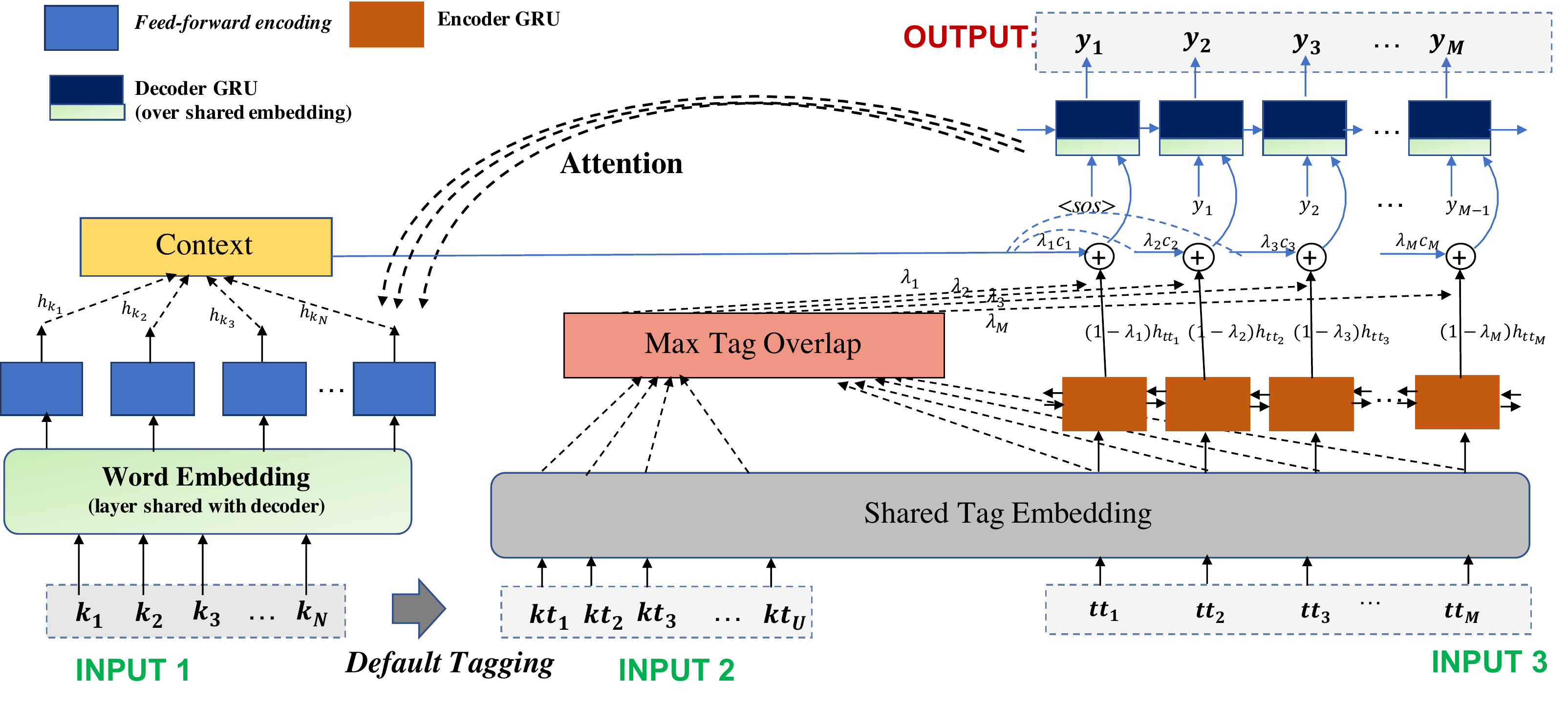}
\caption{Proposed System Architecture}  
\label{fig:system}
\end{figure*}
\section{System Architecture}
\label{sec:system}
For generation of sentences from keywords, the system considers three inputs (padded whenever necessary): (a) a set of $N$ keywords, $K=[k_1,k_2,k_3,...,k_N]$, (b) a set of $U$ \textbf{unique} POS tags
 $KT = [kt_1,kt_2,kt_3,...,kt_U]$ pertaining to the keywords, (c) a sequence of $M$ POS tags, forming the \textbf{template}, $TT=[tt_1,tt_2,...,tt_M]$. The output sentences can be represented as a sequence of $M$ words $Y=[y_1,y_2,y_3,...,y_M]$. Note that, by design, we are keeping the length of the output same as the length of the template. 

The overall architecture is given in Figure \ref{fig:system}. With this input-output configuration, generation is carried out through the modules, as follows:
\subsection{Keyword Encoder}
The objective of the keyword encoder is to capture contextual representations of each keyword in a finite-size vector. It should also ensure that the encoding process is agnostic to the input keyword order. For this, the keywords (in one-hot form) are passed through an embedding layer first. The vocabulary and the embedding layers are shared across the keyword encoder and the decoder. We use a stack of feed forward layers, that non-linearly transform each keyword embedding. Let the transformed vectors be denoted as $H_k=[h_{k_1},h_{k_2},h_{k_3},...,h_{k_N}]$. The decoder extracts appropriate context from these vectors through attention mechanism (explained in Section \ref{sec:decoder}).
\subsection{Template Encoder}
Template encoder is a layer of bidirectional recurrent units (GRUs \cite{chung2014empirical}) stacked on top of an embedding layer of tags. For each time-step of encoding, encoded vectors gathered from both the directions are concatenated and passed through an non-linear layer ($tanh$), on top of MLPs, which reduces its dimension to an appropriate size, as desired by the decoder. Let the encoded templates be represented as $H_{tt}= [h_{tt_1},h_{tt_2},h_{tt_3},...,h_{tt_M}]$.
\subsection{Template and Keyword Tag Matching}
In each step of generation, the decoder has to decide how much contextual information should be considered from the keywords (\textit{i.e.,} from $H_k$) and the template tags (\textit{i.e.,} $H_{tt}$). This is governed by a matching process that yields a probability term  $\lambda$, which is defined as:
\begin{align}
\begin{split}
    s_i = \underset{1 \leq j \leq U}{\max}(cosine(e_{tt_i},e_{kt_j}))\\
    \lambda = sigmoid(W_s^Ts_i+b)
\end{split}
\label{eq:lambda}
\end{align}
Here $\lambda$ is the highest matching probability between the current tag in the template $tt_i$ and one of the tags related to the keywords. $e_{tt_i}$ and $e_{kt_j}$ are embeddings for tags ${tt_i}$ and $kt_j$ respectively. The function $cosine$ is cosine-similarity, commonly used for vector similarity calculation. $\lambda$ helps the decoder decide whether to emphasize on the keyword context or simply ignore the keywords and produce a function word on its own.
\subsection{Decoder}
\label{sec:decoder}
The decoder's job is to construct a probability distribution over the word vocabulary in each time-step. For each time step $t\in\{1,M\}$, the distribution can be given as 
\begin{equation}
    p(y_t|y_1,y_2,...,y_t-1,m_t) = g(y_{t-1},s_t,m_t)
\end{equation}
where $m_t$ is the context extracted from the encoders at time-step $t$, and $s_t$ is decoder's current hidden state and $g$ is a non-linear activation over a linear function (such as $tanh$). In our setting, the context-vector $m_t$ is computed as follows:
\begin{equation}
    m_t = f(\lambda c_t,1-\lambda h_{tt_t})
    \label{eq:component}
\end{equation}
Here $f$ is a non-linear activation function similar to $g$, which considers both keyword and tag contexts in different proportions due to $\lambda$. $c_t$ is a weighted combination of $H_k$, and the weights are computed through  attention mechanism as follows:
\begin{equation}
    c_t = \sum_{j=1}^{N} \alpha_{tj}h_{k_j}
    \label{eq:context}
\end{equation}
\begin{equation}
    \alpha_{tj} = softmax(a(s_t-1,h_{k_j},h_{tt_t}))
    \label{eq:att}
\end{equation}

where, the  function $a$ is a feed forward network used for computing the attention energy \cite{BahdanauCB14}. Note that, for computing attention weights, we also consider $h_{tt_t}$. This ensures that the template tags also influence the attention mechanism and selection of content. In sum, our design allows the decoder to be more flexible in either extracting contextual information from the keywords or the template. It also ensures that unnecessary attention is not given to the keywords if the generation step does not require so.  

It is worth noting that, for calculation of $\lambda$, even though a $max$ operation is involved, due to the use of $sigmoid$, and the choice of similarity function, the network is still differentiable. 
We also observe that when the tag-embedding layers are initialized with unique embeddings for each tag\footnote{We set the tag-embedding dimension same as the tag vocabulary count and using a one-hot vector for each tag during initialization}, the initial learning process becomes more stable. Even though the network initially sees very little similarity across similar tags (e.g., NN and NNS), it gradually brings embeddings of similar POS categories closer as the training progresses. This is desirable as universal POS tags are used for the keywords, whereas fine-grained POS tags are used in the template.

We now describe the experimental setup. 
\section{Experiments}
\label{sec:exp}
\subsection{Dataset}
\label{sec:dataset}
Since our framework requires keywords, tags for the keywords, and tagged template sequences, it is possible to generate a large amount of training data with the help of an off-the-shelf POS tagger and an unlabeled corpus. For English, a large amount of simple sentences are available as a part of the ParaNMT project \cite{wieting-gimpel-2018-paranmt}. We use a derivative of this dataset extracted by \newcite{chen-etal-2019-controllable}\footnote{Obtained from 
https://tinyurl.com/y7rvv4df
}. Sentences in the dataset are tagged using the Spacy\footnote{http://spacy.io/} tagger, and the tagged sequences are retained as template. The sentences themselves are used as gold-standard references. Words of categories NOUN, VERB, ADJECTIVE and ADVERB are \textbf{lemmatized} and used as keywords (ref. INPUT 1 in Figure \ref{fig:system}). For each keyword-set, the POS tagger is independently executed and the unique tags related to all keywords in an example are retained (ref. INPUT 2 in Figure \ref{fig:system}). It is worth noting that POS tagging of a token is context dependent. But since the input in dev/test in our task setting is a list of keywords (instead of a sentence) which could be in any order, tagging for each keyword is performed independently. Table \ref{tab:datastats} mentions the dataset statistics.
\begin{table}
    \centering
    \resizebox{.95\columnwidth}{!} {
        \begin{tabular}{c  c  c  c} 
        \toprule
        \textbf{Split} & \# examples  & Avg. keyword & Avg. sent. length \\ [0.5ex] 
        \midrule
        Train & 959013 & 3.74 & 10.611 \\ 
        Dev & 500 & 2.93 & 8.82 \\
        Test & 800 & 3.411 & 9.58 \\
        \bottomrule
        \end{tabular}}
    \caption{Data statistics. The word vocabulary size used for representing keywords and output sentences is 40738 and the tag vocabulary size is 57.}
    \label{tab:datastats}
\end{table}

For testing, \newcite{chen-etal-2019-controllable} have provided a benchmark dataset which contains an input sentence and an exemplar sentence. From the input sentence, we extract keywords and tags corresponding to the keywords as mentioned above. This creates two possible evaluation scenarios for us: 
\begin{enumerate}
    \item \textbf{\textit{Exact Template}}: evaluated by considering the POS tag sequences of the reference output as template, and 
    \item \textbf{\textit{Similar Template}}: evaluated by considering the POS tag sequences of the exemplar sentences as template.
\end{enumerate}
We observe that the exemplar sentences are often not of same length as the expected output. Using the POS sequences of the exemplars will generate different outputs than the references, making it difficult to evaluate. For a fair evaluation, we report results for both scenarios (1) and (2).
\begin{table*}[t]
    \centering
    \resizebox{2.00\columnwidth}{!}{
    \begin{tabular}{c|c c|c c|c c|c c|c c}
      \toprule
        & \multicolumn{2}{c|}{\bf BLEU (\%)}	&	\multicolumn{2}{c|}{\bf METEOR (\%)}	&	\multicolumn{2}{c|}{\bf ROUGE-L (\%)}	&	\multicolumn{2}{c|}{\bf SkipT}	&	\multicolumn{2}{c}{\bf POSMatch (\%)} \\ \midrule	
\bf Model	& \bf Exact	& \bf	Similar	& \bf	Exact	& \bf	Similar	& \bf	Exact	& \bf	Similar	& \bf	Exact	& \bf	Similar	& \bf	Exact	& \bf	Similar \\ \midrule
\textsc{TransNoTemplate}	& 6.08	&	N/A	&	18.74	&	N/A	&	30.7	&	N/A	&	0.72	&	N/A	&	14.9	&	N/A \\
\textsc{RNNNoTemplate}	&	16.51	&	N/A	&	27.13	&	N/A	&	48.17	&	N/A	&	0.76	&	N/A	&	22.44	&	N/A \\ \midrule
\textsc{TransConcat} & \bf 55.1 & 8.75 & \bf 45.4 & \bf 21.13 & \bf 78.44 & 41.1 & \bf 0.91 & 0.78 & 92.77 & 84.7 \\
\textsc{RNNConcat} & 52.62 & 8.33 & 44.5 & 20.81 & 77.22 & 40.7 & \bf 0.91 & 0.78 & 92.95 & 80.0 \\
\textsc{SentExemp} \cite{chen-etal-2019-controllable} & 36.64 & 2.47 & 29.38 & 12.39 & 65.14 & 30.29 & 0.85 & 0.72 & 67.36 & 53.56 \\
\textsc{Template} (OUR) & 40.33 & \bf 10.12 & 38.15 & 20.96 & 70.46 & 42.77 & 0.89 & \bf 0.79 & 93.08 & 92.15 \\
\textsc{TemplateBeam} (OUR) & 40.79 & 10.0 & 38.28 & 20.94 & 70.66 & \bf 42.79 & 0.89 & \bf 0.79 & \bf 93.03 & 92.16 \\ \bottomrule
    \end{tabular}}
    \caption{Results for sentence generation from input keywords on the dataset by \newcite{chen-etal-2019-controllable}.}
    \label{tab:results-keywords}
\end{table*}

\begin{table*}[t]
    \centering
    \resizebox{2.00\columnwidth}{!}{
    \begin{tabular}{c|c c|c c|c c|c c|c c}
      \toprule
        & \multicolumn{2}{c|}{\bf BLEU (\%)}	&	\multicolumn{2}{c|}{\bf METEOR (\%)}	&	\multicolumn{2}{c|}{\bf ROUGE-L (\%)}	&	\multicolumn{2}{c|}{\bf SkipT}	&	\multicolumn{2}{c}{\bf POSMatch (\%)} \\ \midrule	
\bf Model	& \bf Exact	& \bf	Similar	& \bf	Exact	& \bf	Similar	& \bf	Exact	& \bf	Similar	& \bf	Exact	& \bf	Similar	& \bf	Exact	& \bf	Similar \\ \midrule
\textsc{TransNoTemplate} & 3.75 & N/A & 20.25 & N/A & 26.42 & N/A & 0.68 & N/A &  N/A & N/A \\
\textsc{RNNNoTemplate} & 3.52 & N/A & 20.83 & N/A & 28.12 & N/A & 0.7 & N/A & N/A & N/A \\ \midrule
\textsc{TransConcat} & 18.78 & 7.37 & 28.31 & 20 & 52.27 & 38.9 & 0.82 & 0.77 & 87.27 & 83.04 \\
\textsc{RNNConcat} & 15.89 & 6.25 & 28.07 & 19.6 & 50.68 & 38.01 & 0.83 & 0.77 & 83.87 & 79.93 \\
\textsc{SentExemp} \cite{chen-etal-2019-controllable} & 36.64 & 2.47 & 29.38 & 12.39 & 65.14 & 30.29 & 0.85 & 0.72 & 67.36 & 53.56 \\
\textsc{Template} (OUR) & 40.33 & \bf 10.12 & 38.15 & \bf 20.96 & 70.46 & 42.77 & \bf 0.89 & \bf 0.79 & \bf 93.08 & 92.15 \\
\textsc{TemplateBeam} (OUR) & \bf 40.79 & 10.0 & \bf 38.28 & 20.94 & \bf 70.66 & \bf 42.79 & \bf 0.89 & \bf 0.79 & 93.03 & \bf 92.16 \\ \bottomrule
    \end{tabular}}
    \caption{Results for sentence generation from by \textbf{reversing} input keywords on the dataset by \newcite{chen-etal-2019-controllable}.}
    \label{tab:results-reverse}
\end{table*}

\subsection{Model Configuration}
The word and tag embedding dimensions were set to $500$ and $57$ respectively. The hidden dimensions for decoder, keyword encoder, and tag encoder were set to $500$, $500$ and $100$ respectively. Both encoder and decoder had dropout operations enabled during training with a dropout probability of $0.5$. Cross-entropy loss was considered as the loss criterion and for parameter optimization, Adam optimizer was used with a learning rate of $0.001$. The model trains for $40$ iterations with a batch size set to $256$. Model implementation was done using the \texttt{pytorch} API. 
\subsection{Comparison Systems}
The closest system to ours by \newcite{chen-etal-2019-controllable} (termed \textsc{SentExemp}) is used as baseline. In their original setting, the model expects input and exemplar sentences for training. We retrained the system to consider keyword list and exemplar sentence as input. During testing, for the \textbf{\textit{Exact}} scenario\footnote{The ``Exact'' setup emulates controlled domains where the templates indeed repeat heavily (e.g., QAs in dialogs, dialog acts like greetings, chit-chat, etc.).}, the expected output is given as the exemplar. For the \textbf{\textit{Similar}} scenario, testing is  straightforward as the exemplar sentences are already available in the test dataset. 

Apart from this, we consider four different baselines as mentioned below.
\begin{enumerate}
    \item \textsc{TransNoTemplate}: A transformer based encoder-decoder framework \cite{vaswani2017attention} that only accepts keywords as input and not any template.
    \item \textsc{RNNNoTemplate}: An LSTM based encoder-decoder framework \cite{BahdanauCB14} that only accepts keywords as input and not any template.
    \item \textsc{TransConcat}: Transformer based framework with keywords and templates concatenated and given as input.
    \item \textsc{RNNConcat}: LSTM based framework with keywords and templates concatenated and given as input.
\end{enumerate}

We also consider \underline{\em two variants of our model}: (a) the default (termed \textsc{\bf Template}) model (without beam search), and (b) a variant (termed \textsc{\bf TemplateBeam}) whose best output is obtained through beam search (beam width of $5$) over the output space.

The baseline Transformer and RNN models were trained using the OpenNMT toolkit \cite{opennmt}. All the models were trained with default configurations.
\subsection{Evaluation}
\label{sec:eval}
Through evaluation, we seek to answer the following research questions:
\begin{enumerate}
    \item \textbf{RQ1:} Is the proposed framework able to produce output that is fluent and related to the input keywords? How does the performance fair against baselines and existing systems?
    \item \textbf{RQ2:} Does the structure of output predicted by our framework conform to the specified template?
    \item \textbf{RQ3:} How sensitive is our framework to the variation of specified keywords' order?
    \item \textbf{RQ4:} To what extent can our framework adapt to handle inadequate/spurious information provided in the keyword list? 
\end{enumerate}
\begin{figure}[t]
\centering 
\includegraphics[width=7 cm]{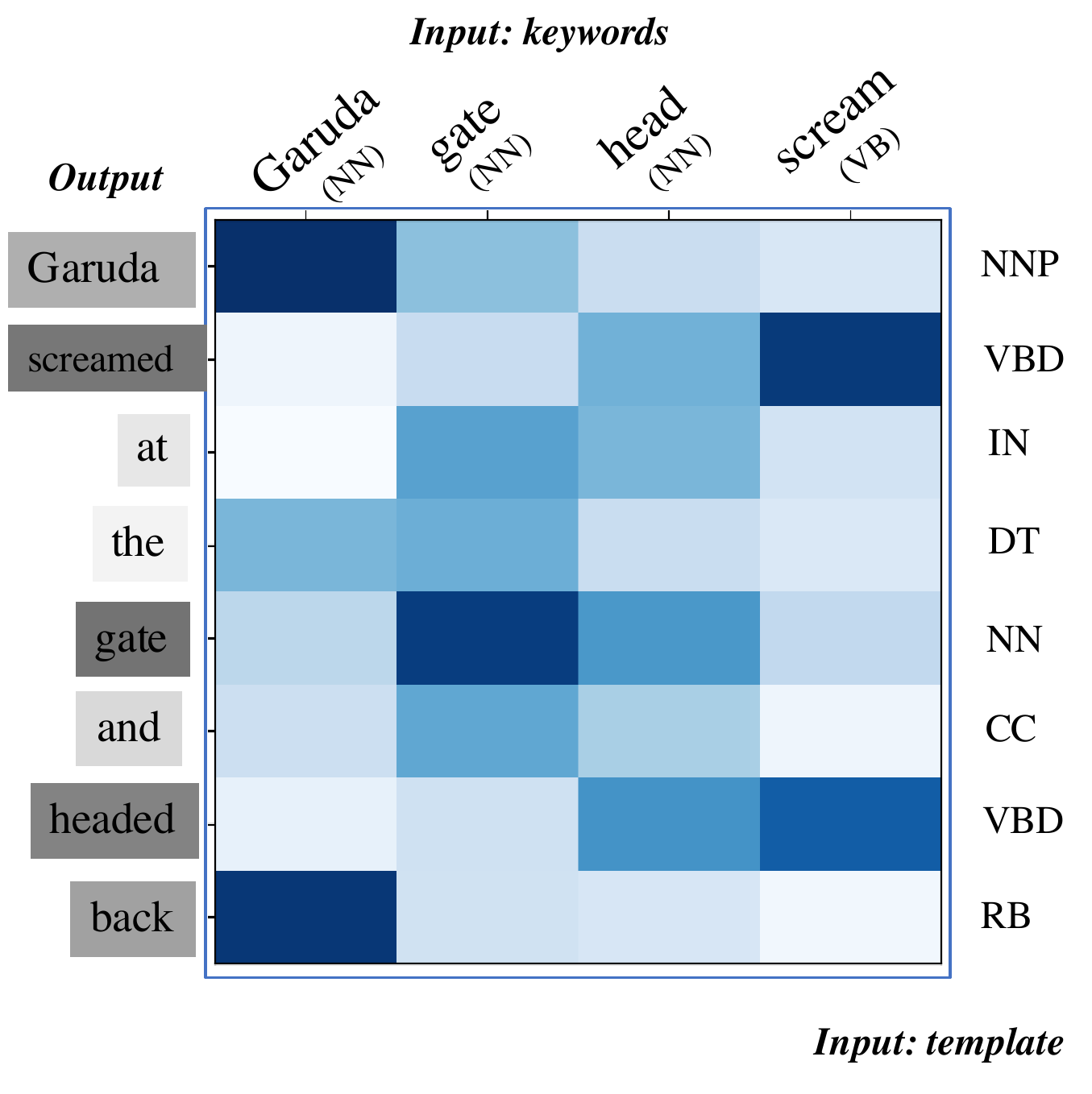}
\caption{Attention heatmap for an anecdotal example. Shades of gray represent $\lambda$ values for the generated words in the output and shades of blue represent attention weights for the keywords. Darker colors correspond to higher values.}  
\label{fig:attn}
\end{figure}
\begin{table*}[t]
    \centering
    \resizebox{2.1\columnwidth}{!}{
    \begin{tabular}{c|c|c|l|c}
      \hline
\bf Keywords and & \bf Example Template & \bf Template POS & \bf Generated Output & \bf Remark \\
\bf Keyword POS & \bf  Sentence &  &  & \\ \hline 
john (NN) , job (NN) & who is bob's  & WP VBZ NNP POS NN . & \textcolor{brown}{\textsc{SentExemp}} who is this right & question form \\  
 & friend? & & \textcolor{orange}{\textsc{RNNConcat}}  what 's john 's job ? & \\ 
& & & \textcolor{blue}{\textsc{TransConcat}} what 's john 's job ? & \\  
& & & \textcolor{red}{\textsc{\bf TemplateBeam}} what 's john 's job ?  & \\ \hline 
john (NN) , job (NN) & why does bob  & WP VBZ NNP VB IN DT NN . & \textcolor{brown}{\textsc{SentExemp}} why does ten go to the boy & paraphrasing with  \\  
 & go to the city? & & \textcolor{orange}{\textsc{RNNConcat}}  what does john do for the job ? & different template \\  
& & & \textcolor{blue}{\textsc{TransConcat}} what does john do about the job ? & \\  
& & & \textcolor{red}{\textsc{\bf TemplateBeam}} what does john do with the job ?  & \\ \hline 
movie (NN) , beautiful (JJ)& this is a horrible  & DT VBZ DT JJ NN & \textcolor{brown}{\textsc{SentExemp}} this is a horrible ...  & declarative form \\  
& story. & & \textcolor{orange}{\textsc{RNNConcat}}  this is a movie beautiful & \\  
& & & \textcolor{blue}{\textsc{TransConcat}} this is a whole movie & \\  
& & & \textcolor{red}{\textsc{\bf TemplateBeam}} this is a beautiful movie & \\ \hline 
movie (NN) , beautiful (JJ) & what a horrible  & WP DT JJ NN PRP VBZ . & \textcolor{brown}{\textsc{SentExemp}} what a movie show this ... & exclamatory form \\  
 & story it is! & & \textcolor{orange}{\textsc{RNNConcat}}  what a whole movie it $<$unk$>$ . & \\  
& & & \textcolor{blue}{\textsc{TransConcat}} what a whole movie it is . & \\  
& & & \textcolor{red}{\textsc{\bf TemplateBeam}} what a beautiful movie it is . & \\ \hline 
movie (NN) , beautiful (JJ)& is it a horrible  & VBZ PRP DT JJ NN . & \textcolor{brown}{\textsc{SentExemp}} is it a horrible ... & interrogative form \\  
 & story? & & \textcolor{orange}{\textsc{RNNConcat}}  is it a movie beautiful ? & \\  
& & & \textcolor{blue}{\textsc{TransConcat}} is it a whole movie ? & \\  
& & & \textcolor{red}{\textsc{\bf TemplateBeam}} is it a beautiful movie ? & \\ \hline 
movie (NN) , beautiful (JJ) & that is not a  & DT VBZ RB DT JJ NN . & \textcolor{brown}{\textsc{SentExemp}} that is not a beautiful ... & negation \\  
 & good book. & & \textcolor{orange}{\textsc{RNNConcat}}  this is n't a movie beautiful . & \\  
& & & \textcolor{blue}{\textsc{TransConcat}} this is n't a whole movie . & \\  
& & & \textcolor{red}{\textsc{\bf TemplateBeam}} this is n't a beautiful movie . & \\  \hline
president (NN) , begin (VB) ,  & the boy started  & DT NN VBD NN . & \textcolor{brown}{\textsc{SentExemp}} the president president 's  & varying keyword  \\  
work (VB) , great (JJ) & crying. & & \textcolor{orange}{\textsc{RNNConcat}} the president began work . & order \\  
 & & & \textcolor{blue}{\textsc{TransConcat}} the president began work . &  \\  
& & & \textcolor{red}{\textsc{\bf TemplateBeam}} the president began work . & \\ \hline 
begin (VB) , work (VB) ,  & the boy started & DT NN VBD NN . & \textcolor{brown}{\textsc{SentExemp}} the president president 's & varying keyword  \\  
president (NN) , great (JJ) & crying. & & \textcolor{orange}{\textsc{RNNConcat}}  the begin worked president . & order  \\  
& & & \textcolor{blue}{\textsc{TransConcat}} the begin worked president . & \\  
& & & \textcolor{red}{\textsc{\bf TemplateBeam}} the president began work .  & \\ \hline 
great (JJ) , president (NN) , & the boy started & DT NN VBD NN . & \textcolor{brown}{\textsc{SentExemp}} the president president 's & varying keyword   \\  
work (VB) , begin (VB) & crying. & & \textcolor{orange}{\textsc{RNNConcat}} the great president work . & order\\  
& & & \textcolor{blue}{\textsc{TransConcat}} the great was president . & \\  
& & & \textcolor{red}{\textsc{\bf TemplateBeam}} the president began work . & \\ \hline 
begin (VB) , work (VB) ,  & the boy started  & DT NN VBD NN . & \textcolor{brown}{\textsc{SentExemp}} the president president 's  & varying keyword  \\  
great (JJ) , president (NN) & crying. & & \textcolor{orange}{\textsc{RNNConcat}}  the begin worked great . & order \\  
& & & \textcolor{blue}{\textsc{TransConcat}} the begin worked great president . & \\  
& & & \textcolor{red}{\textsc{\bf TemplateBeam}} the president began work . & \\ \hline
    \end{tabular}}
    \caption{Examples of generated sentences from different models.}
    \label{tab:example}
\end{table*}
In order to answer these, we first evaluate our models and the baselines with  popular NLG evaluation metrics such as BLEU \cite{papineni2002bleu}, METEOR \cite{banerjee2005meteor} and ROUGE \cite{lin2004rouge}. Additionally, we use Skip-thought sentence similarity metric\footnote{https://github.com/Maluuba/nlg-eval} (termed as \textbf{SkipT}) to check the semantic fidelity between the generated output and reference sentences.

To answer \textbf{RQ2}, we measure the averaged POS overlap (termed as \textbf{POSMatch}). It measures the degree of exact match between the input template and the POS sequence of the predicted sentence. For this, the same POS tagger that generated the templates is applied on the predicted sentences. Percentage of POS match is reported as a part of the results.

To answer \textbf{RQ3}, we reverse the keyword order in the test data and carry out testing of our model variants and baselines on the modified data. Finally, we inspect the $\lambda$ values in Eq. \ref{eq:lambda} and the modified attention weights after multiplying $\lambda$ with the attention weights (Eq.\ref{eq:att}) to answer \textbf{RQ4} and other questions.
\section{Results}
\label{ref:results}
Tables \ref{tab:results-keywords} and \ref{tab:results-reverse} present the evaluation results. Performance indices for the \textsc{NoTemplate} models make it clear that without external knowledge about the syntax and style, it becomes harder even for state-of-the-art sequence-to-sequence models to produce fluent and adequate sentences from keywords. Measuring \textbf{POSMatch} does not help evaluate these models since they do not consider template inputs. The \textsc{Concat} models perform very well when keywords are presented in the same order in which their variations appear in the output, but perform quite poorly when such ordering is not preserved. \textsc{SentExemp} is order agnostic but is not designed for key-word to text generation, thus performs poorly. 
Our \textsc{Template} model variants are stable and provide decent performance in the \textbf{Exact} scenario and are insensitive to change in keyword order. However, they quite strictly follow the template patterns, which reduces their performance when exact templates are not provided. These observations indeed help answer \textbf{RQ1}, \textbf{RQ2} and \textbf{RQ3}.

We present a few examples in Table \ref{tab:example} focusing on different linguistic and practical aspects \textit{i.e.,} variation in syntax and style, change in input keyword order and presence of spurious content in the input. It is evident that our POS based models are capable of handling templates of various sentence forms such as declarative, interrogative, exclamatory and negation. Finally, the last four examples clearly show our model's capability towards ignoring spurious entries (\textit{i.e.} the adjective ``great'') as opposed to the baseline models. This partially addresses \textbf{RQ4}.

Figure \ref{fig:attn} shows the attention weights and $\lambda$ values observed during generation of an example from the test dataset (darker colors indicates higher weights). For constructing this example, four keywords \textit{``Garuda''}, \textit{``gate''}, \textit{``head''} and \textit{``scream''} with default POS categories of \textit{noun (NN)}, \textit{noun (NN)}, \textit{noun (NN)} and \textit{verb (VB)} respectively are provided to our \textsc{Template} model. The input template set to \textit{``NNP VBD IN DT NN CC VBD RB''}, derived from an example compound sentence. It is evident from the figure that content words in the output obtain higher $\lambda$ values and for them the keyword representations extracted through attention mechanism play a vital role. For the other words, the keyword representations are still used, but minimally. Even then, they help in finding appropriate function words (\textit{e.g.} whether to use a definite article or not). Finally, we speculate that by attending over all the keyword representations, the model learned to hallucinate the verb  (``headed'') from the noun (``head'') and applies the correct tense form. This demonstrates that our model has the potential to tackle information gaps in the keyword set, as hypothesized in \textbf{RQ4}.
\section{Conclusion}
\label{sec:conclusion}
We presented a novel weekly supervised approach for text generation from keywords. With the help of templates, and a carefully designed attention mechanism, our system aptly learns to translate keywords into sentences of diverse structure and styles, and also remains unaffected by the change of order in the input keywords. Through our evaluation, we  showed our system's capability in handling inadequate/surplus information in the keywords. And, unlike related previous work, our system simply uses sequences of tags instead of parse trees as templates\footnote{Parsing is known to be considerably more expensive than POS tagging. Also, the accuracy of POS tagging is higher than that of parsing.}.

While the generation quality during test depends on the quality of  template, it is not unreasonable to assume the availability of high-quality templates during run-time. Our observation is that even for the templates used for the ``Similar'' set-up, generated sentences are fluent and adequate.

As a possible follow up work, we would like to use universal POS tags (UPTs)\footnote{https://universaldependencies.org/u/pos/} in templates. The goal is to train our system for keywords in one language as input and generating sentences in another language. This could be feasible since our approach is indifferent to underlying source language or keywords' order. Assessing the model's usefulness in real world data-to-text applications, and for generation of synthetic data for text-to-text systems is also on our future agenda. 


\bibliography{acl2020}
\bibliographystyle{acl_natbib}
\end{document}